\documentclass[10pt]{article}

\usepackage[utf8]{inputenc}

\usepackage{amsmath}
\usepackage{amssymb}

\usepackage{authblk}
\usepackage{comment}
\usepackage{graphicx}
\usepackage{csquotes}
\usepackage{cases}
\renewcommand{\mkbegdispquote}[2]{\itshape}

\usepackage{ccaption}
\usepackage[margin=1.00in]{geometry}

\usepackage{xcolor}

\title{Learning from learning machines: a new generation of AI technology to meet the needs of science}

\author[a,1,2]{Luca Pion-Tonachini}
\author[b,c,d,1]{Kristofer Bouchard}
\author[b,e,f,g,1]{Hector Garcia Martin}
\author[c,h,i,j,k,1]{Sean Peisert}
\author[a,b]{W. Bradley Holtz}
\author[l]{Anil Aswani}
\author[m]{Dipankar Dwivedi}
\author[m,n]{Haruko Wainwright}
\author[o]{Ghanshyam Pilania}
\author[p,k]{Benjamin Nachman}
\author[q]{Babetta L. Marrone}
\author[r]{Nicola Falco}
\author[c]{Prabhat}
\author[s]{Daniel Arnold}
\author[a]{Alejandro Wolf-Yadlin}
\author[t,u]{Sarah Powers}
\author[v]{Sharlee Climer}
\author[a]{Quinn Jackson}
\author[a]{Ty Carlson}
\author[s]{Michael Sohn}
\author[c]{Petrus Zwart}
\author[w]{Neeraj Kumar}
\author[x,y,z]{Amy Justice}
\author[za]{Claire Tomlin}
\author[zb,zc,zd]{Daniel Jacobson}
\author[ze]{Gos Micklem}
\author[b,zf,zg,zh,zi,zj,zk,zl]{Georgios V. Gkoutos}
\author[zm]{Peter J. Bickel}
\author[b,zf,zg,zh]{Jean-Baptiste Cazier}
\author[c]{Juliane M\"uller}
\author[zn]{Bobbie-Jo Webb-Robertson}
\author[zo,zp]{Rick Stevens}
\author[a]{Mark Anderson}
\author[a,zq,1,2]{Ken Kreutz-Delgado}
\author[c,zm,zr,1,2]{Michael W. Mahoney}
\author[b,zm,1,2]{James B. Brown}

\affil[a]{Pattern Computer, Inc., Friday Harbor, WA, USA 98250}
\affil[b]{Biosciences Area, Lawrence Berkeley National Lab, Berkeley, CA, USA 94803}
\affil[c]{Computational Research Division, Lawrence Berkeley National Lab, Berkeley, CA, USA 94720}
\affil[d]{Helen Wils Neuroscience Institute and Redwood Center for Theoretical Neuroscience, UC Berkeley, Berkeley, CA, USA 94720}
\affil[e]{DOE Agile BioFoundry, Lawrence Berkeley National Laboratory, Berkeley, CA, USA 94803}
\affil[f]{Joint BioEnergy Institute, Lawrence Berkeley National Laboratory, Berkeley, CA, USA 94803}
\affil[g]{BCAM, Basque Center for Applied Mathematics, Bilbao, SPAIN 48009}
\affil[h]{Computer Science, University of California, Davis, Davis, CA, USA 95616 }
\affil[i]{CENIC, La Mirada, CA, USA 90638}
\affil[j]{Health Informatics, University of California, Davis School of Medicine, Sacramento, CA, USA 95817 }
\affil[k]{Berkeley Institute for Data Science, University of California, Berkeley, Berkeley, CA, USA 94720}
\affil[l]{Industrial Engineering and Operations Research, University of California, Berkeley, Berkeley, CA, USA 94720}
\affil[m]{Environmental \& Earth Sciences Area, Lawrence Berkeley National Laboratory, Berkeley, CA, USA 94803}
\affil[n]{Nuclear Engineering, University of California, Berkeley, Berkeley, CA, USA 94720 }
\affil[o]{Materials Science and Technology Division, Los Alamos National Laboratory, Los Alamos, NM, USA 87545}
\affil[p]{Physics Division, Lawrence Berkeley National Lab, Berkeley, CA, USA 94720}
\affil[q]{Bioscience Division, Los Alamos National Laboratory, Los Alamos, NM, USA 87545}
\affil[r]{Earth and Environmental Sciences Area, Lawrence Berkeley National Lab, Berkeley, CA, USA 94803}
\affil[s]{Energy Technologies Area, Lawrence Berkeley National Lab, Berkeley, CA, USA 94803}
\affil[t]{Computing and Computational Sciences Directorate, Oak Ridge National Laboratory, Oak Ridge, TN, USA 37831}
\affil[u]{Industrial \& Systems Engineering, The University of Tennessee, Knoxville, TN, USA 37996}
\affil[v]{Department of Computer Science, University of Missouri-Saint Louis, St. Louis, MO, USA 63121}
\affil[w]{Computational Biology Division, Pacific Northwest National Laboratory, Richland, WA, USA 99352}
\affil[x]{Yale University School of Medicine, West Haven, CT, USA 06516}
\affil[y]{Office of Cooperative Research, Yale University School of Public Health, New Haven, CT, USA 06511}
\affil[z]{Connecticut VA Healthcare System, West Haven, CT, USA 06516}
\affil[za]{Electrical Engineering and Computer Sciences, University of California, Berkeley, Berkeley, CA, USA 94720}
\affil[zb]{Energy and Environmental Sciences Directorate, Computer Science and Mathematics Division, Oak Ridge National Laboratory, Oak Ridge, TN, USA 37831}
\affil[zc]{The Bredesen Center for Interdisciplinary Research and Graduate Education, University of Tennessee, Knoxville, Knoxville, TN, USA 37996}
\affil[zd]{Department of Psychology, University of Tennessee Knoxville, University of Tennessee, Knoxville, Knoxville, TN, USA 37996}
\affil[ze]{Department of Genetics, University of Cambridge, Cambridge, UK CB2 3EH}
\affil[zf]{Centre for Computational Biology, University of Birmingham, Birmingham, UK B15 2TT}
\affil[zg]{Institute of Cancer and Genomic Sciences, University of Birmingham, Birmingham, UK B15 2TT}
\affil[zh]{The Alan Turing Institute, London, UK NW1 2DB}
\affil[zi]{Institute of Translational Medicine, University Hospital Birmingham NHS Foundation Trust, West Midlands, UK B15 2GW}
\affil[zj]{MRC Health Data Research UK (HDR),  Midlands Site, UK}
\affil[zk]{NIHR Surgical Reconstruction and Microbiology Research Centre, Birmingham, UK B15 2TT}
\affil[zl]{NIHR Biomedical Research Centre, B15 2TT, Birmingham, UK}
\affil[zm]{Department of Statistics, University of California, Berkeley, Berkeley, CA, USA 94720}
\affil[zn]{Biological Sciences Division, Pacific Northwest National Laboratory, Richland, WA, USA 99352}
\affil[zo]{Computing, Environment, and Life Sciences, Argonne National Laboratory, Lemont, IL USA 60439}
\affil[zp]{Department of Computer Science, University of Chicago, Chicago, IL USA 60637}
\affil[zq]{Calit2/Qualcomm Institute, Pattern Recognition Laboratory, University of California San Diego, La Jolla, CA, USA 92093}
\affil[zr]{International Computer Science Institute, University of California, Berkeley, CA, USA 94704}
\affil[1]{Contributed equally to this manuscript}
\affil[2]{Corresponding authors: Luca Pion-Tonachini, luca@patterncomputer.com; Ken Kreutz-Delgado, kenkd@patterncomputer.com; Michael W. Mahoney, mmahoney@stat.berkeley.edu; James B. Brown, jbbrown@lbl.gov}

\begin{document}

\maketitle


\newpage

\begin{abstract}
We outline emerging opportunities and challenges to enhance the utility of AI for scientific discovery. 
The distinct goals of AI for industry versus the goals of AI for science create tension between identifying patterns in data versus discovering patterns in the world from data. 
If we address the fundamental challenges associated with ``bridging the gap'' between domain-driven scientific models and data-driven AI learning machines, then we expect that these AI models can transform hypothesis generation, scientific discovery, and the scientific process itself.
\end{abstract}


\section{Introduction}




Today, in science and engineering, the means by which we obtain hypotheses and models is taken for granted: it is human intuition---informed and constrained by data and deep domain knowledge.
In contrast, in Machine Learning (ML) and Artificial Intelligence (AI), one uses learning machines, e.g., Support Vector Machines, Deep Neural Networks, and other related data-driven AI models.
These methods enable astounding predictive performance in a relatively domain-agnostic manner in computer vision, natural language processing, and certain related areas.
Despite these remarkable predictive successes and its increasing impact on science, AI has not yet had the kind of transformative impact on science that it has had on society more generally.%
\footnote{We frame much of our discussion in terms of AI, but it could equivalently be framed in terms of ML, as there is little consensus on the terms and domain boundaries in this area.}
More generally, we have seen AI alter the fabric of society, with the creation of new industries, markets, and business models, leading to seismic shifts in our democracy and the democratic process itself. 
Not all of these changes are necessarily welcome, but they are profound.
They are indicative of the arrival of a technology as momentous as calculus, the steam engine, statistics, and the electronic computer. 
The question before us is ``Will science be as changed and as unrecognizable after the widespread integration of AI as it was after the arrival and adoption of the electronic computer?'' 
We believe so.
Our goal here is to outline some of the fundamental challenges in AI research that are needed to accelerate this transformation. 

There have been and will continue to be predictive successes for AI in science:
in biology, AI algorithms are used to predict biomolecular structures and to guide the bioengineering process \cite{jumper2021highly,carbonell2019opportunities, radivojevic2020machine, zhang2020combining, streich2020can};
in materials science, we are beginning to see benefit from AI-designed materials \cite{tshitoyan2019unsupervised};
in climate science, AI is used to reduce uncertainty in the prediction of future climate scenarios \cite{gentine2018could};
in cosmology, AI is used to conduct simulations of unprecedented scale of the early universe \cite{he2019learning}; 
in neuroscience, AI is critical to processing massive volumes of anatomical connectomics data \cite{hpcbrain2018}; and 
AI is increasingly used for challenges such as drug design \cite{gawehn2016} and the management of cultivated ecosystems in precision agriculture \cite{CHLINGARYAN201861,harfouche2019accelerating}.
More interestingly, decision-support utilities and self-driving labs \cite{macleod2020self}---wherein hypotheses are formulated, tested, and refined by AI agents---may herald a coming revolution. 
The most important feature of the self-driving lab paradigm is not the acceleration of the tedious processes of discovery.
It is the translation of insights obtained from data by an AI model into scientifically interpretable hypotheses that can be tested---and ultimately understood. 
These nascent systems aim to do more than predict \emph{what} will happen, they attempt to offer insight into \emph{how} or~\emph{why}.

The disproportionate impact of AI on society versus AI on science stems, in large part, from differences between goals in science---which typically aim to understand natural or engineered phenomena and processes in the world---and goals in industry---which typically aim to solve specific problems and achieve positive return-on-business-investment.
These differences lead to different guiding theories and practices, different questions being asked of AI models, differences in whether AI models are treated in a black-box or white-box manner, and differences in whether AI models are simply ``reproducing the phenomena'' or whether they are capturing properties of the world.
As a result of this, many AI methods develop by industry predict \emph{what} will happen, but they offer little insight into \emph{how} or \emph{why}.
Obtaining this insight is the goal of science.
The gap between understanding \emph{what} happens and understanding \emph{how} and \emph{why} it happens is a major impediment to realizing the full potential of AI in many scientific domains.
\emph{To deliver on the promise of AI for science, we need a science of AI: we must develop methods to extract novel, testable hypotheses directly from data-driven AI models.}
These AI-enabled hypotheses can then be tested in the same way that any other scientific hypothesis is tested.

In this overview, we outline emerging opportunities and challenges to enhance the utility of AI for scientific discovery. 
The distinct goals of AI for industry versus the goals of AI for science create tension between identifying patterns in data versus discovering patterns in the world from data. 
If we address the fundamental challenges associated with ``bridging the gap'' between domain-driven scientific models and data-driven AI learning machines, then we expect that these AI models can transform hypothesis generation, scientific discovery, and the scientific process itself.

\section{Scientific first-principles models and data-driven statistical or AI models}

In science, we aim to understand the world.
Thus, given the increasing importance of AI models in the scientific process, we begin with a question: 
\begin{displayquote}
	What does it mean to ``understand'' a model or to ``obtain understanding from'' a model---AI or~otherwise?
\end{displayquote}
The answer to this question depends on who we ask and how they were trained.

For many data scientists and ML practitioners, ``understanding'' can only be given meaning in terms of the ability of a model to make high-quality predictions on unseen data.
In this case, there is a methodology that AI model developers typically adopt: split the data into two (or three, if a validation set is used) groups; train on the first group; measure performance on the second group; adjust model architecture and hyperparameter values in light of the results; and iterate \cite{bishop2006pattern, murphy2012machine, Goodfellow-et-al-2016}. 
This type of approach---which tries to obtain high-quality prediction on the data that have been generated, but which typically does not engage in counterfactual reasoning about that data---is the primary means by which AI models are developed, trained, and deployed in many of the most high-profile industrial applications.


Statisticians and statistical data scientists and practicing domain scientists, however, tend to have somewhat different views.
Statisticians rarely, if ever, think of statistical models or AI models (e.g., Support Vector Machines, Random forests, Boosted Ensembles, and Deep Neural Networks) as representing first-principles, in the sense usually attributed by practicing domain scientists to scientific models. 
George Box famously summarized this way of thinking when he said ``All models are wrong, but some are useful'' \cite{box71p}, which
Bickel and Doksum heralded as the ``guiding principle of modern statistics'' \cite{bickel2015mathematical}. 
Practicing domain scientists, however, would say that Newtonian mechanics is ``wrong'' in a very different way than Ptolemaic and Copernican mechanics were ``wrong.''
The latter involved fitting epicycles, and is regarded as a canonical example of ``curve fitting,'' which in its time yielded excellent predictions, but little understanding of why the world functions as it does.
Newtonian mechanics involved a minimum of assumptions and a large body of falsifiable predictions, and has been shown to remain valid well beyond its original domain of applicability. 
It is regarded as a canonical example of a first-principles scientific model, grounded in a few deep principles, inductively arrived at from data, from which deep understanding about the world is obtained.


For statisticians, understanding arises from the interrogation of fitted models, i.e., not simply evaluating models by the quality of their predictions on training/testing data, but by examining the models themselves---both for understanding the models themselves, and for understanding the world that those models are modeling.
This interrogation typically (but not necessarily, as we discuss below) involves hypothetical or counterfactual reasoning applied to data representations, i.e., to models themselves and not just to model outputs, and it provides a way to discern when a statistical model is \emph{not} appropriate to use.
This interrogation also often points toward hypotheses about mechanisms---mechanisms which, by their nature, also may well remain valid well beyond the data used for their discovery. 
Statistical interrogation techniques include inference, hypothesis testing, uncertainty quantification, and the derivation of confidence regions---all foundational procedures throughout data science. 
For example, to assess the impacts of parameters or covariates in linear models, we often make use of confidence regions, Bayes factors, \textit{p}-values, \textit{q}-values, and measures of effect size. 
We are not advocating for or against these measures, and there are important subtleties in their use that are often ignored in practice.
We are simply pointing to their existence and, for better or for worse, the practical purposes they serve in scientific research, as it is practiced today.
In more complex statistical models, non-parametric approaches such as permutation tests, stability analysis, and other measures serve similar roles \cite{yu2018, Yu3920, Breiman2001, hooker2019please}. 
Importantly, each interrogation method used to obtain such understanding from a fitted model is dependent on the ability to go beyond model predictions, to access directly and to reason quantitatively about the data representation constructed during model fitting.

\begin{figure*}[p]
\begin{center}
\includegraphics[width=0.74\textwidth]{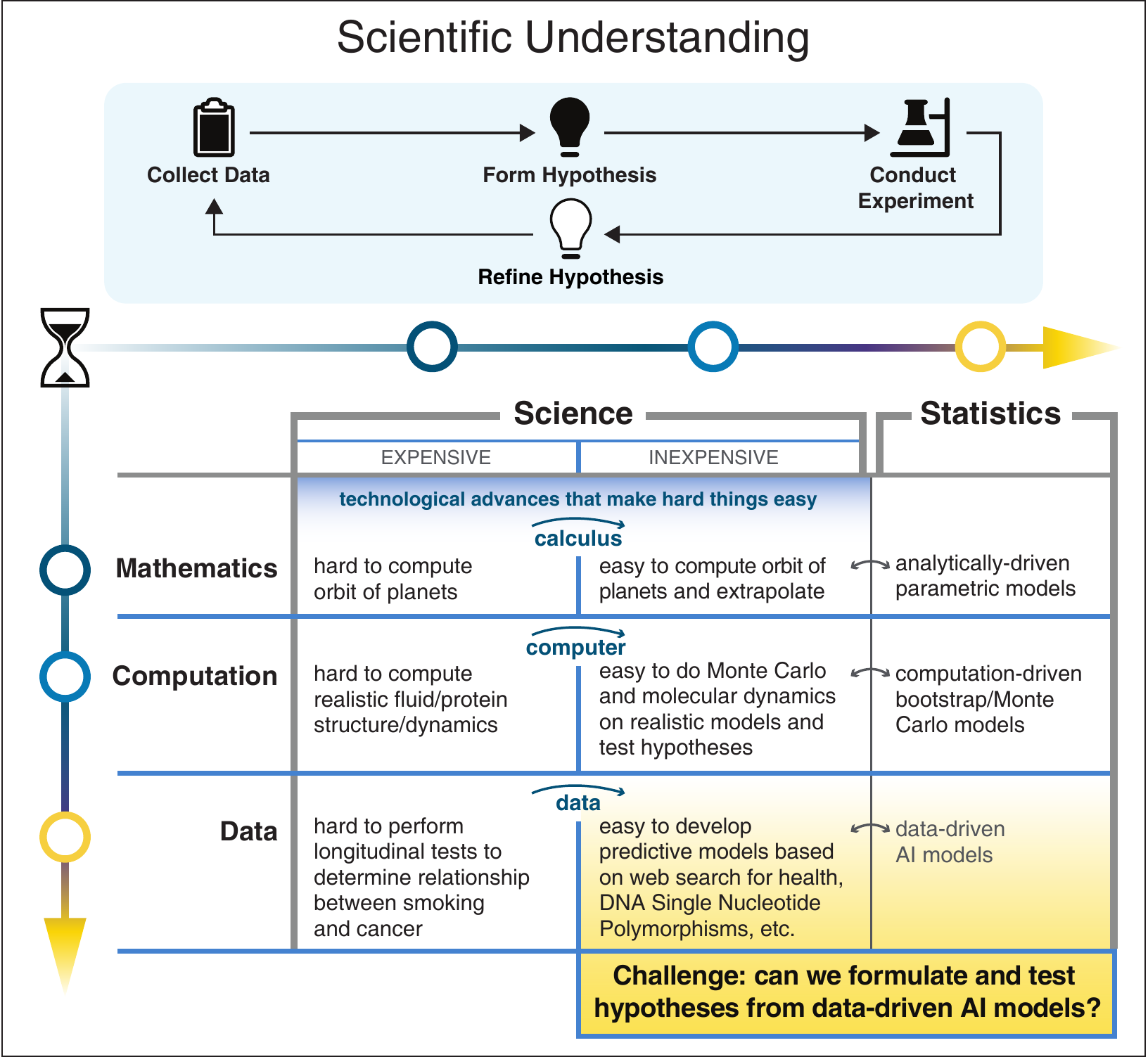}
\caption{Overview of cumulative improvement of tools and technologies. 
Top: the cycle of scientific discovery---where statistical methods and/or counterfactual reasoning and hypothesis testing are used to refine hypotheses and, ultimately, yield understanding and knowledge. 
Bottom: the timelines of innovation in scientific and statistical tools for hypothesis discovery and refinement. 
Change-points in our capacity for understanding the world are highlighted, from top to bottom and left to right: 
the innovation of calculus that led to the (inexpensive) computability of celestial mechanics; 
the arrival of electronic computers and associated numerical methods that led to modeling complex and realistic scientific systems; and, most recently, 
new data generation instruments and associated advanced computation that have resulted in the aggregation of data on transformative scales.
In parallel with these foundational advancements in the scientific method, we have advancements in statistics (rightmost column)---new tools for modeling the processes by which we can learn about the world. 
The dawn of statistical reasoning formalized the process of hypothesis testing, and it provided a language with which to discuss confidence, uncertainty, and measures of association. 
Computers enabled the modeling of far more complex processes with statistical rigor---nonparametric statistics gave rise to powerful ML methods, such as Random Forests. 
Now, in the era of vast datasets, data-driven AI technologies (bottom right), first designed in the previous century, enable the modeling of systems of previously intractable complexity. 
The challenge before us concerns the extent to which these data-driven AI technologies can co-mingle with statistical reasoning to transform the scientific method as thoroughly and completely as did calculus and the electronic computer.
%
}
\label{fig:history}
\end{center}
\end{figure*}

For practicing domain scientists, understanding also arises from the interrogation of fitted models, albeit in a somewhat different way.
The scientific method provides a well-established approach for humans to learn from data: observe the world by collecting data, form hypotheses and make falsifiable predictions, test these predictions by collecting new data via controlled experiments, and iterate.
This approach, as well as related scientific methods applied to observational data, also provides a way to know when a scientific model is \emph{not} a reliable phenomenological representation.
Throughout the years, decades, and centuries, many different tools---analytical, computational, and data-driven---have been and continue to be used to aid this process.
Figure~\ref{fig:history} provides an overview.
Within the last few centuries, analytical theory and other mathematical tools have been used to provide clarity and precision. 
Within the last few decades, computational tools have become available to extend dramatically the capabilities of pencil-and-paper human ``computers'' to work with more complex and realistic models.
Within the last few years, data-driven tools themselves are increasingly used as integral components of the process of scientific discovery.
Regardless of whether analytical, computational, or data-driven tools dominate the discussion, to obtain scientific insight, these tools must be used within the context of the scientific method to formulate and evaluate testable hypotheses with controlled experiments.
This methodology is very mature for analytical theory and computational techniques \cite{agrawal2016paradigm}, but it is much less developed when it comes to using data-driven AI models to obtain scientific~insight.

In both cases, statistical data science and domain science, obtaining understanding about the world from a model requires going beyond making high-quality predictions on unseen data to asking counterfactual statistical questions and/or making and evaluating falsifiable predictions.
In the same way that traditional statistical models can be analyzed during the process of inference (deciding which model features are likely to be significant and estimating parameter values) or uncertainty quantification (measuring trust in learned effect sizes), we look to future state-of-the-art AI models for the same capabilities.

Statisticians often begin this process by attempting to describe a ``minimal model'' that explains a system's behavior in terms of as small a collection of parameters as possible (under some regularized objective, e.g., via the Lasso, Bayes Information Criterion, etc.). 
Seeking such a minimal model can produce hypotheses that relate tractable collections of parameters to outcomes---and the twin processes of inference and uncertainty quantification can be used to rank or prioritize those hypotheses for subsequent experimentation as part of the cycle of scientific discovery. 
The property of minimality is at the core of first-principle models: it aims to describe system behavior in terms of testable mechanisms and nothing else. 
Accurate mechanistic or physical models are intrinsically minimal---they include only the parameters that modulate system behavior: the principles from which $\textbf{F}=\textit{m}\textbf{a}$ can be derived contain no extraneous or redundant information or~parameters.

\begin{figure*}[p]
\begin{center}
\includegraphics[trim=70 320 70 55,clip,width=\textwidth]{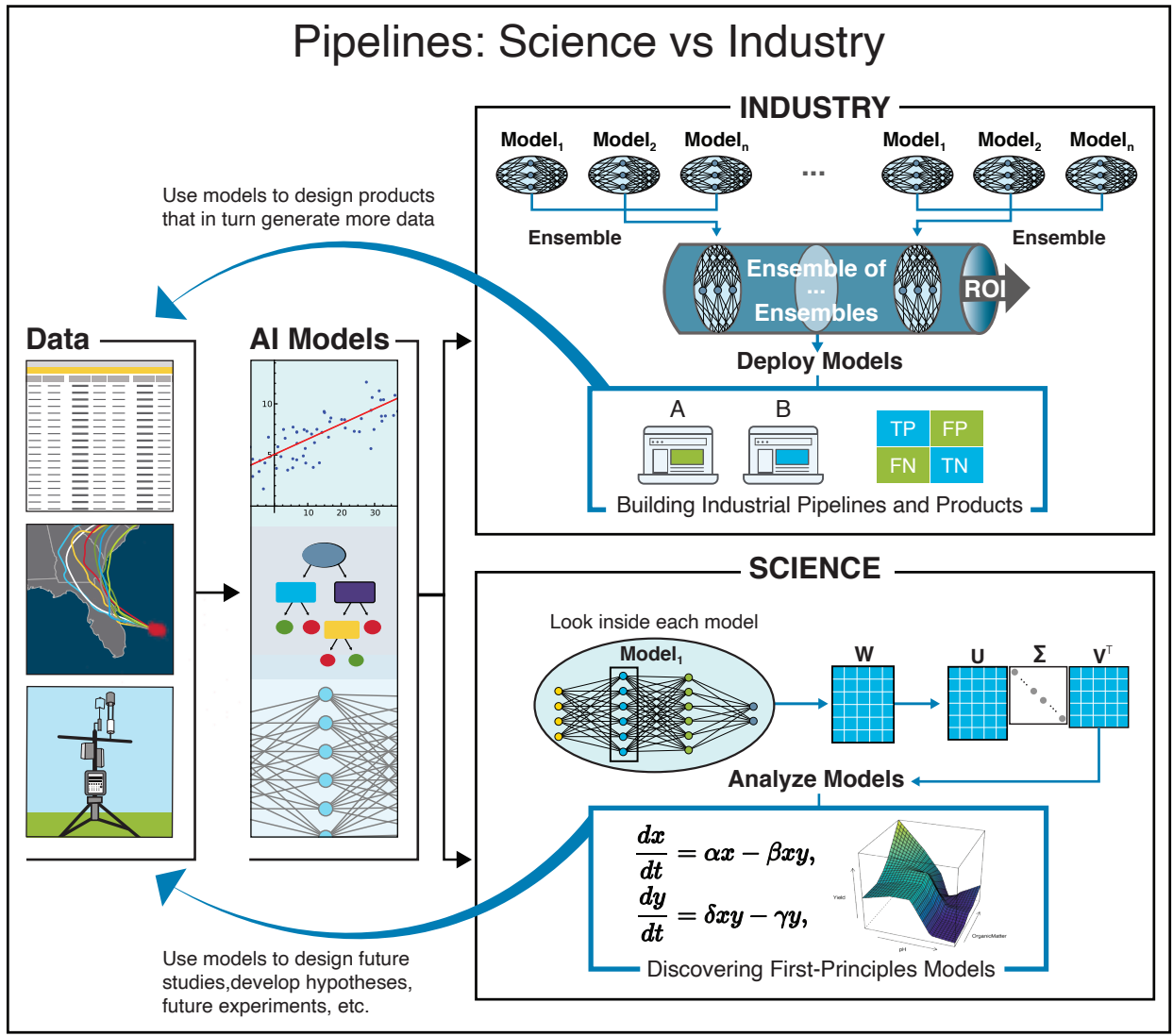}
\caption{Scientific versus industry applications of state-of-the-art AI.  Both involve data collection and model fitting, but the intended use of those data-driven AI models is fundamentally different: scientific models aim to enable understanding of the world; while industry models must perform well for industry goals. How these models iteratively interact with data is also very different: the use of scientific models leads to new hypotheses, which leads to new experiments, which leads to new data; while the use of industry models leads to products that generate more data, which in turn leads to the development of models that are more performant, which leads to more data.  Much of the recent innovation in state-of-the-art AI methodology has been driven by industry goals.}
\label{fig:pipelines}
\end{center}
\end{figure*}

While there are important differences between recently popular AI methods and traditional statistical methods, popular AI learning machines share far more in common with traditional statistical models than with first-principles models from domain science \cite{pearl2019seven,FP19_TR}.
This is in large part because of disparate scientific versus industry applications of state-of-the-art AI and with the differing goals of science versus industry; see Figure~\ref{fig:pipelines} for an overview.
This creates a challenge for domain scientists---e.g., biologists, chemists, physicists---who want to extract knowledge from AI models that have been fit to data from that scientific domain. 
How then can AI models be extended to incorporate the same kinds of interrogation that drives hypothesis discovery and guides hypothesis testing and experimental design in domain sciences? 
To answer this question, we begin with a discussion to build some common vocabulary across disciplines.


\section{Generalization versus extrapolation}

Achieving scientific understanding from data-driven AI models embedded in a scientific workflow 
requires access to data and data representations in order to discover patterns---not just in the data, but ultimately in the world.
This consists of at least two components:
\begin{enumerate}
    \item 
    The capacity to learn models from data that lead to testable hypotheses---potentially about data that are very different than the original data used to train the models. 
    \item 
    The capacity to recognize and learn from ``interesting'' outliers---which might simply be errors or which might be the harbinger of new science.
\end{enumerate}
Importantly, while both components are central to the scientific method, neither is a priority \emph{per se} for current-generation AI algorithms, which have developed principally outside the realm of science, in internet, social media, and related areas of industry.%
\footnote{By ``industry,'' we tend to mean internet, social media, and related industries that have driven much of the recent developments in AI methods.  Certainly, other industries are increasingly using AI methods.  We expect that the approaches we advocate will be particularly relevant for them.}
For example, AI models are used in industry to build deployable systems, e.g., an image classification system or a text sentiment analysis system or an ad placement system, but those systems are not probed, as a matter of course, to formulate and test hypotheses about why the AI system works, or which patterns it has learned.  
An AI model in the context of these industrial applications is useful if it performs well and if it can be integrated into an automated system that leads to a competitive advantage (e.g., as measured by profit-based metrics).

Similarly, there are always outliers, but in industry outliers in the data are usually not of interest in and of themselves.
They typically represent data to be cleaned or risks to be hedged, e.g., to avoid poor system performance or to avoid bad publicity. 
In contrast, in science such outliers may constitute the first hint of unanticipated observations and new scientific insights, e.g., that a theory is in some way insufficient or overreaching and should be reconsidered. 
For instance, deviations in Newtonian predictions has led to insights about the world: deviations in the orbit of Uranus led to the discovery of Neptune; and deviations in the orbit of Mercury let to the confirmation of general relativity.
The treatment of outliers, and what we attempt to learn from them, differs fundamentally between many industrial applications and typical scientific use cases.

\begin{figure*}[p]
\begin{center}
\includegraphics[width=0.95\textwidth]{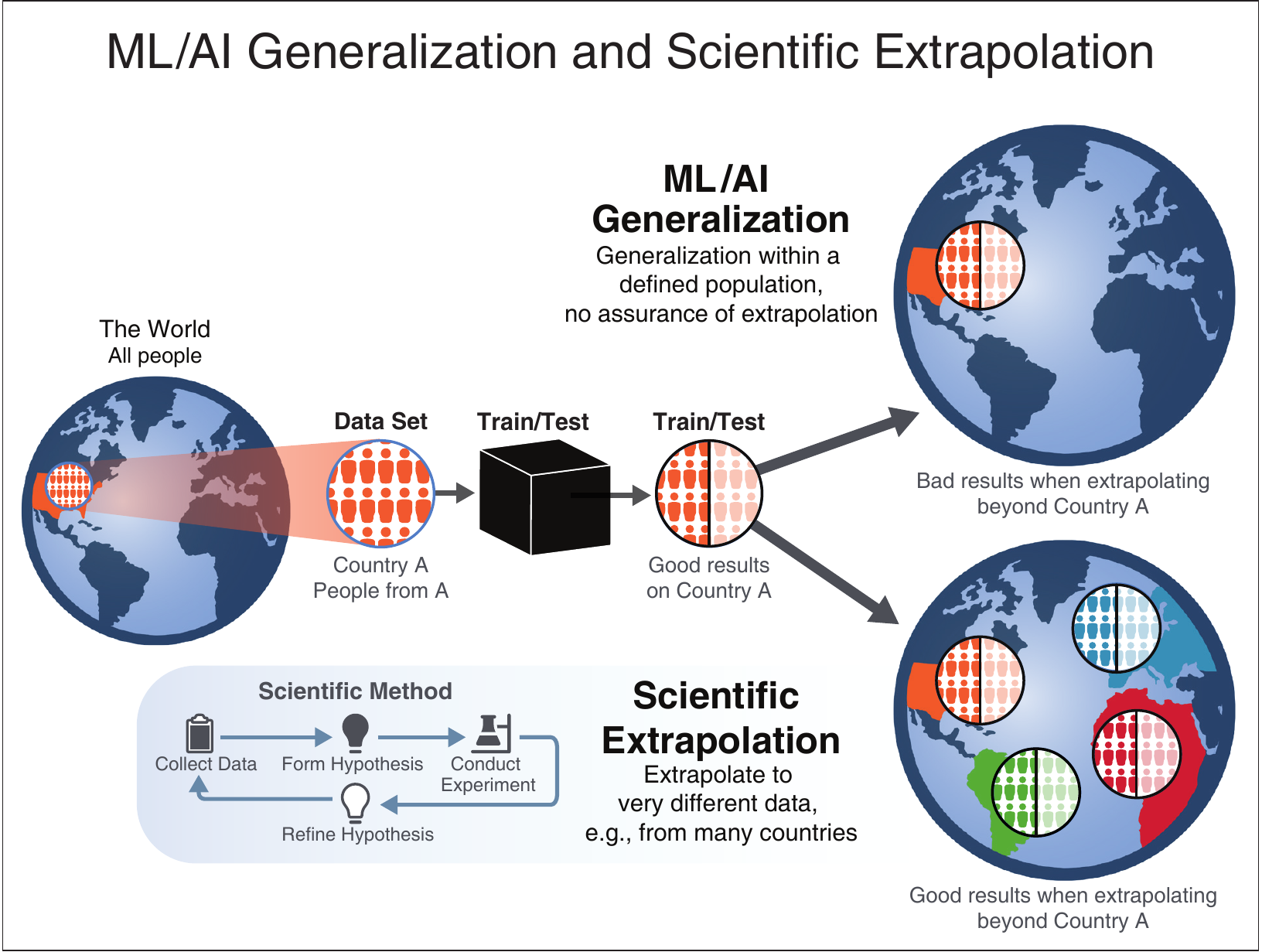}
\caption{Generalization versus extrapolation. 
Given a model trained on a subset or subpopulation of a larger set or total population, how that models performs can be characterized in at least two ways: generalization and extrapolation. 
Generalization (top right) refers to the model's ability to perform well on data that are equivalent (up to random training-testing splits) to the data on which it was trained. 
Extrapolation (bottom right) refers to the model's ability to perform well on data that are fundamentally different than the data on which it was trained (e.g., data from different subpopulations than the training set). 
The minimization of ``generalization error'' is the standard by which the predictive accuracy of AI algorithms is often (i.e., almost always) judged, and it is achieved through fitting methodologies (e.g., gradient assent, bagging, cross validation) that help the model learn properties of the training data. 
Extrapolation, the transferability of a model to data that is fundamentally different than that on which it was trained, is achieved through scientific counterfactual methodologies, which help the model to learn properties of the world.}
\label{fig:generalization_versus_extrapolation}
\end{center}
\end{figure*}

Scientists expect a good model---both first-principles models as well as phenomenological or semi-empirical models built upon first-principles models---to be applicable well outside the narrow domain of the data used for model construction and fitting. In science, this notion is known as extrapolation.
In AI research, this often corresponds to what is called transfer learning~\cite{utrera20a_TR}.
This notion is critically different than (even if it is colloquially similar to) the more narrow notion of generalization~\cite{hast-tibs-fried} (which is central to ML theory underlying AI methods).
Figure~\ref{fig:generalization_versus_extrapolation} illustrates this difference.

Generalization is a key theoretical concept that drives the development of AI models \cite{bishop2006pattern,murphy2012machine}.
Generalization accuracy is usually assessed by splitting a given data set into two or more pieces, e.g., a training set and a testing set; and training an AI model on one half and testing or evaluating (with precision, recall, etc.) that model on the other half.
Thus, model fitting consists of optimizing performance on data that are the ``same'' (in expectation, or on average, i.e., up to the training-testing split) as the data on which the model was fitted.
How an AI model performs on a very ``different'' dataset, or how it performs on data that could have been generated but were not, does not enter into either the construction or the evaluation of the~model.

Extrapolation is a very different notion.
The assessment of a model's capacity for extrapolation requires the use of counterfactual reasoning, e.g., statistical hypothesis testing or controlled scientific experiments, to make claims about data that were not, or could not, be collected.
Essentially, the difference between generalization and extrapolation concerns whether AI models are discovering patterns within the data, or discovering patterns from the data that extrapolate to the world, and then to new data generated from different scenarios. 
The example of Copernican mechanics versus Newtonian mechanics is apt here: both fitted models are capable of performing well, but fitted epicycles failed spectacularly when new data were generated (unless still more and more epicycles were added), whereas the principles of Newtonian mechanics extrapolated broadly---far beyond the domain in which they were discovered.

In the current absence of a statistical or scientific methodology to evaluate, validate, and understand AI models, these models---when applied outside the narrow domain of the dataset on which they were trained, which is the typical use case---may extrapolate well, but they may extrapolate very poorly.
There is no guarantee either way, and there is little practical theory to guide users of these models.
Alternately, they may also be brittle and lack robustness, leading to serious and even catastrophic errors. 
This has long been known, e.g., in robust statistics and in many domain sciences, but it has received renewed attention recently even within AI research.
For example, this lack of robustness lies at the root of the existence of adversarial attacks on fitted AI models \cite{AM18_adversarial}, where one can construct imperceptible (to the human) perturbations of natural images that are classified completely differently, and other forms of adversarial AI \cite{ganin2016domain}.
For industrial AI, this lack of robustness is a major practical concern.
For scientific AI, this matters even more, as AI models increasingly act as the lens through which we view data.
A lack of robustness affects the veracity and scope of conclusions we draw from models about the world.

\section{Explainability versus interpretability}

Visually compelling examples of these issues have been highlighted recently by the ``explainable AI'' community, or XAI as DARPA has coined the term \cite{samek2019explainable}, which has focused on a few key areas: saliency maps \cite{Adebayo2018}; relevance attribution \cite{MONTAVON2018,sundararajan2017axiomatic}; filter selection and compression \cite{abbasiasl2017interpreting}, etc.
These methods identify and rank regions of the data (by highlighting pixels, usually in images or video segments in computer vision use cases) that have a large impact on the model's final decision. 
Related methods that have been proposed to advance the application of AI models in science include ``Intensive PCA'' \cite{Quinn2019}, which maps input-output relationships in Neural Networks to an explorable space, akin to PCA, and Uniform Manifold Approximation and Projection (UMAP) \cite{McInnes2018}, which has been used to extract structures between inputs and outputs of Neural Networks \cite{carter2019activation}. 
As with other methods of regression diagnostics~\cite{ChatterjeeHadiPrice00}, these methods can be useful: they may aid in exploratory data analysis; and they may reduce the workload of the human in the loop \cite{reyes2020interpretability}, e.g., when examining many radiological images or a huge number of astronomical images.
However, these methods are developed by determining properties of the data responsible for good prediction accuracy, e.g., pixels with large gradient values during the training process---not via scientific or statistical counterfactual analysis.
Thus, the ``explanation'' is fundamentally about the model and the internal mechanics of the model, and not about the world and properties of the world that generated the data.

Said another way, a key difference between AI for science and AI for industry, as outlined in Figure~\ref{fig:pipelines}, has to do with ``explainability'' of AI models versus ``interpretability'' of AI models.
These two terms are often used interchangeably and inconsistently.
More important than the specific terms is the underlying concepts they try to capture.
While superficially similar, the underlying concepts are fundamentally very different.
The difference lies in the distinction between identifying patterns in the data/model (hence, ML-style generalization) versus discovering patterns in the world from data/model (hence, scientific extrapolation, or ML-style transfer learning).
Explainability, as with XAI more generally, concerns how well the internal mechanics of a specific AI model can be explained in human terms, i.e., \emph{how} an AI model works. 
Interpretability concerns how well properties of the world, e.g., cause and effect, can be observed and discovered and understood, when using that AI model as a lens on the the natural or artificial system that generated that data, i.e., \emph{why} an AI model works, in terms of properties of the world.
Much of the work that is branded as ``explainable AI'' does not---and fundamentally cannot---provide this latter sort of description, except incidentally.
Hence, explainability in itself is insufficient. To endow a model with both explainability and interpretability, as is commonly of interest in science, one must go beyond the specific data and specific model to engage in counterfactual reasoning. 
Importantly, this is how scientists vet models---the process by which a ``model'' becomes or evolves into a~``theory.''

Partly in response to these and related shortcomings, some of the AI community has suggested that it may be desirable to decouple feature importance from representation learning \cite{ribeiro2016should,raffin2019decoupling,stooke2020decoupling}. 
For scientific inquiry, however, this decoupling is only useful if the result is human-comprehensible and interpretable (as defined herein). 
If model inputs have semantic meaning in terms of the domain from which the data are drawn, and if we are interested principally or exclusively in their marginal effects\footnote{Here, the term ``marginal'' is used in the statistical sense, e.g., the impacts of individual model parameters on global properties such as the overall averages or variances.}, then such measures of importance may be of preeminent value. 
As when leading eigenvectors can be interpreted in terms of properties of the domain from which the data are drawn, however, this fortuitous situation arises not just due to the model, but also due to how the data are generated and preprocessed~\cite{Gould96,CUR_PNAS,DKM20_CSSP_neurips}.
If, on the other hand, individual features have strong effects only on finer-scale local properties, as in the context of complex or higher-order dependencies, then such representations may be of little value.

Consider the case of precision medicine, where one is often interested in understanding the relationship between an individual's genetics and their likely response to specific therapeutic interventions \cite{sagner2017p4}. 
When genes function independently, or additively, then marginal effects are often sufficient to formulate counterfactual hypotheses about system behavior.
However, when gene functions depend in detail on broader genetic background, as in so-called complex traits (e.g., susceptibility to cardiovascular disease) \cite{huang2012epistasis}, then it becomes essential to model gene interactions (and, often, also gene-environment interactions) to predict phenotypes~\cite{behr2020learning}. In this setting, the capacity of a model to extrapolate beyond the population on which it is trained depends on its completeness, particularly with respect to higher-order genetic and environmental interactions, that may manifest only in specific subpopulations \cite{bustamante2011genomics}. 

The same reasoning applies to the identification of interesting outliers, where rigorous uncertainty quantification is essential if one wants to obtain scientific insight. 
In this setting, we need to understand the space in which data have been embedded (typically implicitly) by an AI model. 
How we define outliers and how we measure their significance depends in detail on the metric we use to assess the ``distance'' between observations, e.g., between a new observations and those present in the training set \cite{li2018metric}. In addition to surprising individual observations, it is also important to consider ``outliers in density''---collections of observations that, taken together, constitute an unexpected preponderance of data in a given region, e.g., with respect to past observations or an underlying model. The assessment of outliers in density requires density estimation---and therefore an embedding, just as with individual observations.  
Traditional statistical models such as linear regression models deal with the embedding problem directly and explicitly. 
Under a linear model fitted by ordinary least squares (which induces a Euclidean metric structure on the data), it is straightforward to identify data points that should be viewed with caution, or interest---outliers. 
Indeed, statistical measures of leverage, influence, and confidence are available to inform and quantify our uncertainty \cite{ChatterjeeHadiPrice00}.
For much more complex models, such as state-of-the-art Neural Networks, the structure imposed on the data by the model is far less obvious, and traditional metrics may not be appropriate. 
These challenges constitute a barrier to the integration of AI algorithms with scientific pipelines---a barrier we must overcome if we are to have trustworthy measures of confidence and uncertainty for AI models. 

In spite of all of these challenges, it is increasingly clear that AI models can be capable of such extrapolation and of enabling novel scientific insight---potentially in ways quite different than traditional theoretical and computational methodologies. 
AlphaFold provides perhaps the most prominent example \cite{callaway2020will,jumper2021highly,townshend2021geometric}.
More generally, 
recent work on Deep Hidden Physics models enables the extrapolation of system dynamics directly from observations \cite{raissi2018}.
Similarly, in some areas of climate and environmental and materials science, AI models out-perform state-of-the-art physics-based models designed by domain scientists~\cite{RCSx19,rutz2019atmospheric,aditi_persistent}. 
These and other results are strongly suggestive of the potential for data-driven AI methodologies to transform scientific inquiry and even science itself, potentially in a way analogous to and as transformative as the electronic computer did over the last few generations.
To deliver on this promise, however, the components, principles, and interactions captured by these AI models, and how they interact with broader domain-specific methodologies and constraints, must be understood better.
These models currently remain largely opaque, in part since they are remarkably complex, but also since they have thus far been driven by generalization and industrial goals, rather than extrapolation and scientific goals.
The capacity to understand data representations and principles learned by AI models constitutes a tantalizing prospect---and grand challenge---with the potential for significant impacts on scientific discovery and the origination of novel theory.

\section{Learning representations and decision functions}

There are by now examples of AI methods discovering physical conservation laws and dynamics from observational data alone.
For example, recent work \cite{liu2020ai} used symbolic regression on a small number of semantically meaningful covariates to discover equations that describe system behavior. 
In this example, all variables had \emph{semantic meaning}. 
A more challenging class of problems involve features that lack intrinsic semantic meaning---think image or video data, or scientific data in which the semantics are less well-understood. 
In this setting, semantics must be discovered---which is itself a scientific goal.

To illustrate this distinction, imagine attempting to learn the ideal gas law, given only the trajectories of a very large number of particles over time. 
This system has a vast number of degrees of freedom, and it is not clear what are the primitive scientific concepts.
Can an AI algorithm discover emergent parameters like Pressure and Temperature that determine the behavior of the system at equilibrium? 
Can it learn $PV=nRT$ from the observation of particle trajectories alone? 
This problem is non-trivial because the input data, particle trajectories in this case, have little or no direct semantic relationship to the ``control'' variables, $P$ and $V$ in this case, and they must be discovered---along with the physics that govern their~relationships. 

The recent success of AlphaFold in predicting protein structure from primary amino acid sequence has been hailed as one of the most important accomplishments in science in the past half century \cite{callaway2020will,jumper2021highly,townshend2021geometric}. 
The relationships between primary sequence and 3D molecular structure are non-obvious, hence the significance of the problem and achievement. 
This work built on a large body of domain science, but it also involved the use of non-trivial AI methodology.
Thus, in some sense, these AI algorithms are already learning ``emergent'' system features not previously known.
We can ask: can we learn new principles that govern protein structure through the interrogation of the AlphaFold model? 
Does the AlphaFold model have hidden within it knowledge about protein structure that humans, at present, lack?  
If so, how can we extract that knowledge in a scientifically meaningful non-anecdotal way?

It is helpful to consider another example where there is a clear need to discover as-yet-unknown system controls. 
In neuroscience, we can now observe neurons firing together and individually \cite{vladimirov2014light}, and we can simulate the behavior of large groups of neurons arranged into super-structures like cortical columns \cite{schwalger2017towards}. 
However, we do not yet know the correct ``resolution'' at which to study (real, not artificial) neural information processing in living systems as complex as vertebrate brains.  
Should we be studying individual neurons, cortical columns, or other levels of spatial organization? 
How does lateral information transfer within a cortical layer integrate brain regions? 
Can the interrogation of successful predictive models, AI or otherwise, help to crack the codes through which brains compute \cite{einevoll2019scientific}? 

Here, too, there is not a clear connection, if any, between the input data and the scientific control parameters and primitive scientific concepts.
In each of these examples, we lack knowledge of the relevant scale at which the system should be studied---scale has to be learned, along with the semantics and governing principles relevant to that (or those) scale(s). 

To make these distinctions concrete, recall that an AI model can be viewed as a function $f$ that takes as input some data $x$ and returns as output a decision or response $y$.
Then, $f$ can be viewed as a composition of two functions,

\begin{equation}
\label{eqn:f_eq_h_comp_g}
y = f(x) \qquad \text{with} \qquad f = h \circ g \qquad \text{and} \qquad g : \mathcal M \rightarrow \mathcal N ,
\end{equation}

\noindent 
where $f$ is the overall model, $g$ computes the model's data representation, $h$ computes the decision/response, and the composition $h \circ g$ indicates that $g(\cdot)$ is applied to $x$, and then $h(\cdot)$ is applied to $g(x)$.
The sets $\mathcal M$ and $\mathcal N$ can be subspaces, other subsets of high-dimensional Euclidean spaces, or other metric or ultra-metric spaces.
Often, but not always, it holds that $\dim(\mathcal N) \ll \dim(\mathcal M)$.
The function $g$ typically depends on many parameters and hyperparameters that are adjusted during the model fitting process.
It is well-known in AI that ``representations are important'' and that different AI models $f$ and different decompositions of a given AI model $f$ into components $g$ and $h$ provide different, complementary views on the data~\cite{Goodfellow-et-al-2016}. 
In this way, $g$ encodes the patterns learned by the AI model, and $h$ encodes the map that links these patterns to outcomes in the world, $y$. 
For example, with a $20$-layer Feed-Forward Neural Network, it may be useful to choose $g$ to consist of all layers except the last decision layer; or it may be useful to choose $g$ to consist of initial convolutional layers, and to study the data representation computed up to that point; or it may be useful to further disaggregate both $h$ and $g$ more finely.
Of course, the formal decomposition given in Eqn.~(\ref{eqn:f_eq_h_comp_g}) is always possible, as $g$ can be trivially chosen to be the identity, in which case $h = f$.
Table~\ref{tab:f_eq_h_comp_g_exs} provides several non-trivial natural examples.
What concerns us here is not the representations themselves, but instead how the representations are evaluated and used in the broader scientific pipeline.

\begin{table*}[t] 
    \centering
    \begin{tabular}{ p{40mm} | p{60mm} | p{45mm} }
    \hline
    $f = h \circ g$ & $g$ & $h$ \\
    \hline
    \hline
    Linear Regression (LR) \newline \hspace{5mm} with $\ell_1$ regularization           
       & sparse binary vector along with a dot product that specifies the support 
       & weighted sum of the non-zero support \\
    \hline
    Support Vector Machine \newline \hspace{5mm} (SVM)  for classification      
       & kernel function 
       & learned separating hyperplane \\
    \hline
    Random Forest (RF)              
       & ensemble of estimates across all splits in trees prior to final splits that give rise to leaf nodes 
       & ensemblization across leaf nodes \\
    \hline
    Feed-forward Neural Network (NN)
       & all layers of the network prior to the final 
       & final prediction layer \\
    \hline
    \end{tabular}
    \caption{Several popular AI models $f$, and one example for each decomposition of $f$ into $f = h \circ g$.}
    \label{tab:f_eq_h_comp_g_exs}
\end{table*}

Assume that we have obtained an AI model $f$, decomposed as $f = h \circ g$, that achieves good predictive performance on a specific task.
For example, it may predict whether an individual is sick or healthy, or whether a star is old or young, or whether a protein folds into a given 3D structure. 
How such a model is used in industry differs significantly from how such a model is used in science.
Recall that Figure~\ref{fig:pipelines} outlines these two very different approaches.


If one wants to use $f$ to build automated decision systems to be used in an industrial or commercial system, then a common \emph{modus operandi} is the following: develop many such models, each of which is slightly different and makes slightly different errors; and then use a set of techniques known as ensemble methods to combine these models into a single model that is of higher predictive quality than any single model~\cite{bishop2006pattern,dietterich2000ensembles,Halevy_unreasonable_data,bell2007netflix}.
These may then be put into industrial pipelines optimized to deliver business value, and they can be continuously improved with new data as they are deployed. Conformal prediction techniques now make it possible to obtain accurate prediction intervals (measures of uncertainty) in such settings under only the assumption of access to statistically exchangeable observations \cite{bates2021distribution}. 
In this case, where we only need to know \textit{what} an AI model $f$ has learned from data, it is often sufficient to access and analyze $f$ in a black-box manner, and it is often accepted that this model is completely non-understandable---neither explainable nor interpretable.
Even when one has access to a decomposition of $f$ into $g$ and $h$, one typically does not engage in counterfactual reasoning because there is no perceived need. 
That is, one rarely \emph{interrogates} the model to discover or test underlying, potentially causal,~relationships.

If, on the other hand, one wants to use $f$ to obtain scientific insight, then
simply training $f$ and examining the predictions of $f$ are not sufficient.
We require steps after model fitting to extract the drivers, controls, and mechanisms that give rise to system behaviors.
We need to embed AI model development into the life cycle of scientific discovery, in a form that enables counterfactual reasoning, and the generation and testing of hypotheses about mechanisms. 
To accomplish this, we need to understand representations and parameterizations discovered by $g$, as well as their meaning, encoded by $h$, in terms of principles from domain science. 
The structure of representations learned in $g$ can reveal potentially causal factors, and the action of $h$ on these factors maps their relationships to outcomes.

\section{Structure extraction from learned representations
}

To bridge the gap between current, state-of-the-art data-driven AI models and first-principle scientific models, we need to understand (scientifically) the types of structures that AI models can identify in data. 
As an example, when working with statistical models based on well-defined objective functions, the statistics community has 
aimed to discover patterns and build domain understanding by enforcing sparsity, e.g., via regularization in those models \cite{hast-tibs-fried,wainwright2019high}. 
This regularization may be explicit, as with LASSO-based methods~\cite{Tib96}, or it may be implicit, as with CUR-like methods~\cite{CUR_PNAS,Paschou07b,DKM20_CSSP_neurips}.
Low-rankness, sparsity, low-rank plus sparse---these are all common modeling assumptions.
Conversely, when working with models based on less well-defined objectives, techniques have emerged that aim to extract from those models features that are explicitly low-dimensional, e.g., filter selection \cite{lecun1989,hassibi1993,han2015deep} or model order reduction and distillation \cite{hinton2015distilling}. 
These techniques use (typically non-counterfactual) methodologies to determine important parts of an already-trained model; and they also proceed from the presumption that models with fewer parameters, e.g., minimal models, are sufficient to explain or understand the behavior of the systems they approximate. 
In general, traditional statistical theory suggests that whenever it is possible to fit accurate predictive models from finite datasets, it must be possible to describe the system in terms of some sort of low-dimensional representations~\cite{Levina2005,Bickel:1050479}.

This makes sense, as truly high-dimensional systems are challenging to work with: 
high-dimensional dynamical systems tend to exhibit chaotic behavior \cite{Ispolatov2015, Tomlin2003, Aswani2009}; 
distances and angles concentrate in high-dimensional spaces \cite{aggarwal_curse_of_dim}; 
expander graphs exhibit poor isoperimetric and inferential properties \cite{HLW06_expanders}; and 
small amounts of noisy connections short-circuit fine-scale local structure in informatics graphs \cite{LLDM09_communities_IM}. 
Similarly, it is challenging when model complexity and quantity/quality of data diverge together: 
linear regression when the number of data points $n$ and number of features $p$ are large and comparable \cite{ZCM20_TR}; 
informatics graphs when the number of edges is large and comparable to the number of nodes \cite{Jeub15}; 
probabilistic graphical models in the thermodynamic limit \cite{MezardMontanari09}; and
state-of-the-art AI models in computer vision and natural language processing with a huge number of parameters trained on large quantities of data \cite{BHMM19}.
In aggregate, we take these findings to indicate two things: 
first, that an AI model that exhibits good predictive accuracy (in particular when coupled with good extrapolation behavior) must be learning, at least approximately, perhaps very approximately, some sort of parsimonious model; and 
second, that existing methodological machinery (including both statistical approaches and the traditional scientific method) is typically insufficient to identify and learn from that parsimonious model in a scientifically useful way.

Overall, we see the trajectory of innovation in statistics and AI modeling as an arc proceeding from simple, well-defined models (which conform to our \textit{a priori} prejudice that interpretable models should be explicitly and globally low-dimensional, depending on only a small number of parameters) towards increasingly complex models (which may not be analyzable with traditional statistical theory).
These more complex models may be intrinsically, but not explicitly, locally low-dimensional.
Alternatively, these more complex models may, in some sense, couple locally low-dimensional structures into larger more complex intermediate-scale structures, that themselves do \emph{not} aggregate well into a global structure that is easy to analyze. 

A taxonomy of models and structures, visualized in Figure~\ref{fig:global-intermediate-local}A and B, is~helpful.

\begin{itemize}
    \item[] 
    \textit{\textbf{Global Sparsity}}: 
    A model that exhibits \emph{exact global sparsity} depends only on a small subset of all possible parameters. 
    Examples include sparse linear regression and sparse generalized linear models (e.g., with $\ell_0$ or $\ell_1$ regularization). 
    Explicitly interpretable statistical models usually fall into this category \cite{buhlmann2011statistics, CUR_PNAS, bouchard2017union, DKM20_CSSP_neurips}, as do exactly low-rank models, stochastic blockmodels with a constant number of clusters, etc.
    Many other models exhibit \emph{approximate global sparsity}, in the sense that the model depends primarily on a small subset of its parameters but also weakly on many other parameters. 
    We think of these models as perturbative approximations of globally sparse models.
    Ridge regression, which shrinks but does not zero-out small effects, provides a canonical example. 
    Other examples include ridge-regularized low-rank models, mean-field stochastic blockmodels, and manifold-based methods that have been popular in~ML.
    \item[] 
    \textit{\textbf{Local Sparsity}}: 
    A model that exhibits \emph{exact local sparsity} can be viewed as consisting of piecewise low-dimensional models. 
    Here, for any sufficiently small neighborhood around an observation, the model is low-dimensional, meaning that all variation is confined to a relatively small ``intrinsic dimension'' within a (potentially) much higher-dimensional ``ambient'' space. 
    Importantly, however, such models may lack global sparsity, as described above, especially when there exist sufficiently many locally sparse neighborhoods. 
    Cancer biology provides a good example:  nearly 4\% of human genes have been implicated as causal mutations across cancer types \cite{sondka2018cosmic}---yet, in any given patient, only a handful of genes are relevant \cite{sondka2018cosmic,futreal2004census}. 
    Hence, an interpretable model of oncogenesis will be locally sparse, but not globally sparse, with a total intrinsic dimension $\gtrsim 700$ (4\% of human genes).
    Multivariate Adaptive Regression Splines provide an example of a model explicitly designed to exhibit exact local sparsity \cite{Friedman1991}; most boosted models and models based on fitting decision trees fall into this category as well; as do models implicitly-defined by strongly local spectral methods \cite{MOV12_JMLR}. 
    For many other models, which exhibit \emph{approximate local sparsity}, for any sufficiently small neighborhood around an observation, the model is approximately low-dimensional, in some meaningful sense. 
    \item[]
    \textit{\textbf{Intermediate-scale structure:}}
    A model of data has \emph{intermediate-scale structure} if it cannot meaningfully or usefully be formulated as a perturbative approximation of a purely local model or a purely global model.
    For example, a model that exhibits \emph{approximate} local sparsity may also exhibit non-trivial intermediate-scale structure that might also be of interest.
    Recent theoretical and empirical work suggests that state-of-the-art Neural Networks, i.e., the AI models implicitly defined by state-of-the-art deep learning procedures, fall into this category \cite{MM18_TR_JMLRversion,MM20a_trends_NatComm}. 
    These models have millions or billions of parameters, and they learn representations of data that are---at least very approximately---locally low-dimensional.
    However, these representations are then composed in complex ways that lead both to very high predictive quality as well as to recalcitrance to traditional global statistical theory and local data modeling methods. 
\end{itemize}

\noindent
The key point is that for models with intermediate-scale structure, around any given observation, the predictive model may admit a low-dimensional representation that is nearly equivalent in both its learned representation, $g$, and the predictive performance of its decision function, $h$; while at the same time these locally low-dimensional representations do \emph{not} aggregate meaningfully into a low-dimensional representation of the form $f(x) = h \circ g$ for the entire dataset. 
For example, if we view a traditional model with exact global sparsity as function
$f(x) = h(g(x_{i_1},\ldots,x_{i_n}))$, 
with $i_1,\ldots,i_n \in \{ 1,\ldots,N \}$ and $1 \le n \ll N$, 
then we can formalize this notion of locality as: 
%
%
\begin{equation} 
\label{eq:sparsity}
f(x) =
\begin{cases}
h_{1}(g_{1}(x)); & x\in U_{1}\subset\mathbb{X} \\
\qquad\vdots & \qquad\vdots \\
h_{m}(g_{m}(x)); & x\in U_{m}\subset\mathbb{X}
\end{cases}
\end{equation}

\noindent 
with $x \in \mathbb{X}$ and $g_i:\mathbb{X} \rightarrow \mathbb{R}^{n_i}$ and $1\leq n_{1},\dots,n_{m} \ll N$, where we have $N$ features for each observation and the subsets $U_{i}$ define non-overlapping neighborhoods in $\mathbb X$, the domain of $f$ (though lower-dimensional projections may overlap). 
This model is local in the sense that each neighborhood $U_{i}$ defines a sub-model $h_{i}(g_{i}(x))$ that specifies the behavior of $f(x)$ in this region; however, the properties of the data and/or how the model (and hyperparameters, training protocols, etc.) interacts with the data mean that this information does not aggregate into a model with approximate global sparsity.

\begin{figure*}[p]
    \centering
    \includegraphics[height=0.9\textheight]{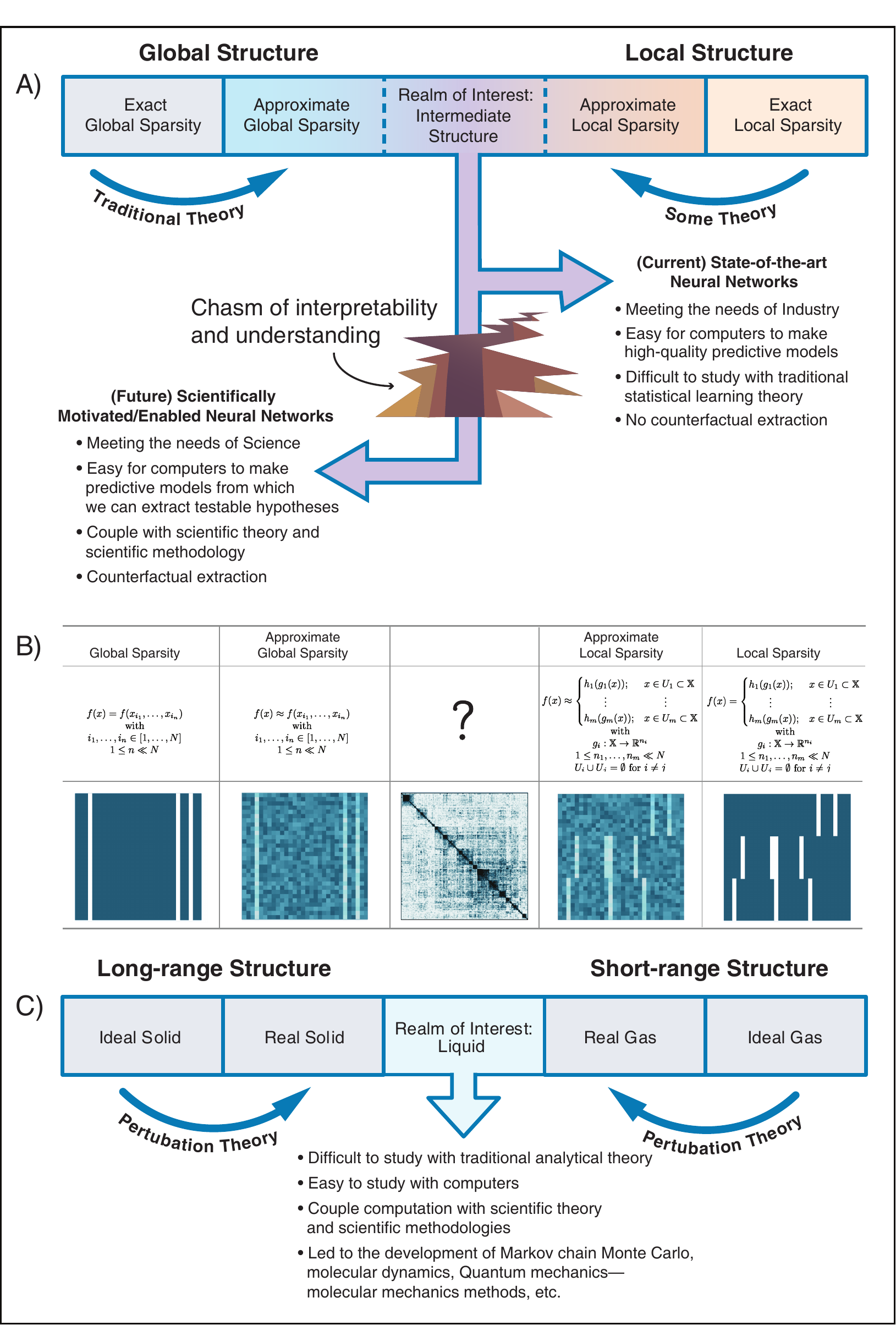}
    \caption{Intermediate scale structure in state-of-the-art AI models, and an historical analogy to physics. 
    (\textbf{A}) Structures that AI models can identify in data.}
    \label{fig:global-intermediate-local}
\end{figure*}

\begin{figure}[!t]
    \centering
    \contcaption{Traditional statistical models (e.g., linear regression, generalized linear or additive models), as well as some early ML models (e.g., support vector machines, stochastic block models), enforce exact or approximate global sparsity (top left). 
    Decision trees, random forests, and a variety of boosted ensembles may lack explicit penalized regularization; yet they produce models that are either exactly or approximately locally sparse (top right). 
    The theoretical properties of these and other local (or locally-biased) methods is not nearly as well understood as that of more common global methods, but we are seeing the beginnings of theory \cite{MOV12_JMLR,biau2019neural}. 
    By contrast, we understand vanishingly little about the types of structures learned by state-of-the-art AI models. 
    (\textbf{B}) The top row of B illustrates examples of formal definitions of global and local sparsity; and observe that it remains unclear how best to define the types of intermediate-scale structures learned by state-of-the-art AI models \cite{MM18_TR_JMLRversion,MM20a_trends_NatComm}. 
    The lower row of B illustrates the types of support learned by global and local and more realistic state-of-the-art AI models. 
    The abscissa corresponds to the parameters of model, with lighter (white) pixels indicating important (non-zero) parameters; and the ordinate corresponds to the observations used to train the model---or alternatively to observations in held-out test data used to evaluate the model. 
    (\textbf{C}) Historical analogy to physics and the electronic computer. 
    Analytical theory was successful at predicting the behavior of matter in solid and gas phases, but it failed to perform well for liquids.
    New computational methodologies were developed to simulate the behavior of liquids, and eventually other more complex states of matter.
    This led to computationally-driven surrogate models that approximate otherwise intractable physics and that are now widely used in science and engineering.  
    }
\end{figure}

Most statistical theory, in particular counterfactual methods for inference, hypothesis testing, and confidence intervals, focuses on models with exact or approximate global sparsity.
Methods to identify exact or approximate local sparsity can sometimes also be analyzed with similar theory, subject to being ``locally-biased'' in some way~\cite{MOV12_JMLR,LBM16_JRNL}.
However, the results for these locally-biased models tend to be much more brittle, since the locally-low dimensional structures are embedded in much higher-dimensional ambient spaces.
The intermediate-scale regime is particularly relevant for extracting insight about the world from AI models---precisely because multiple layers, each consisting of element-wise nonlinear transformations of the data, enable one to extract the local, then intermediate-scale, then more global features that provide high-quality prediction ability \cite{MM18_TR_JMLRversion,MM20a_trends_NatComm}.
If all models are wrong but some are useful, then of course there may be no need for taxonomizing models with this intermediate-scale category.
The same holds true if we are simply interested in evaluating the predictive success of a model and obtaining high-quality prediction/generalization or other forms of non-scientific understanding. 
However, if we aim to extract domain insight from our models, using statistical and/or scientific counterfactual methods to enable extrapolation, then it is restrictive to work only with models that can be treated with existing analytical theory as perturbative approximations of models with global or local sparsity structure.

To understand (scientifically, as we used the term) an AI model requires having the ability to explore the drivers of its predictive performance---to extract local structures $f_j(x) = h_j(g_j(x))$, for $j \in \{1,\ldots,m\}$ (such as those we visualize in Figure~\ref{fig:global-intermediate-local}A and B), and to ask counterfactual questions of the local models $f_j$ and/or their decompositions into $h_j$ and $g_j$. 
Importantly, this can certainly be done when treating $f$ globally and/or as a black box.
Many systems studied by scientists are black box in this sense.
However, in that case, one should only hope to obtain very coarse insight, akin to what is revealed by the leading singular vectors of data matrices, or to the main effects in an additive statistical model.
When one is interested in finer-scale insight, as with the analysis of the drivers of complex genetic traits of interest in precision medicine, working with local models, $f_j$ and their decompositions, $h_j$ and $g_j$, is essential.

The grand challenge here involves going beyond working with purely global or purely local models, where traditional theoretical and computational methodologies are well-developed.
It requires understanding how local models interact, and learning what AI models have learned about the intermediate-scale regime, where local structures couple together in complex, data-driven ways to enable high-quality prediction.
This is the regime where traditional analytical theory and traditional computational approaches perform poorly; but it is also the regime where data-plus-computation have led to state-of-the-art AI models, especially in computer vision and natural language processing applications.
It is a grand challenge since these models enable very high-quality prediction, and thus good generalization, but they do not necessarily enable good extrapolation, e.g., since they can be very brittle to data perturbation or model misspecification.
They do not currently stand up to the scrutiny of counterfactual analysis, which involves reasoning about data that were not observed, and hypotheses about properties of the world that have not been seen.
If we address this grand challenge, then we can go beyond traditional statistical counterfactual techniques---widely-used for global sparsity and local sparsity---to develop scientific counterfactual techniques appropriate for intermediate-scale structure in AI models that are not meaningfully perturbative approximations of purely global models or local models.

\section{Data-driven surrogates to extract scientific insight
}

To address this grand challenge, we need what we will call ``surrogate models,'' in particular scientifically-validated or counterfactually-validated surrogate models.
In our view, an explanation that is useful for scientific or non-scientific goals is itself a model, albeit a simpler one. 
Such a ``surrogate model'' is simply a model of a model. 
A surrogate model may be simpler than it's corresponding full model in the sense that it is smaller and computations on it are less expensive.
Examples of such computationally-motivated surrogates include coresets in computational geometry \cite{AHV06} and sketches in randomized numerical linear algebra \cite{DM16_CACM}.
Alternatively, a surrogate model may be easier to understand, reason about, or perform counterfactual analysis upon, even if this means sacrificing ideals such as rigorous theory or accepting reduced predictive~performance. 

\emph{We propose that the development of scientifically-validated surrogates for state-of-the-art AI models is needed to meet the needs of data-driven or AI-enabled science.}
If AI models provide a ``lens'' through which we view data, then these surrogate models will enable us to discover what these AI models have learned. 
Scientifically-validated surrogates are much more than a smaller model that closely matches the AI model's learned responses on a given dataset. 
Matching a response is not equivalent to matching the mechanism by which that response was achieved.
This is the essential point when considering the potential for extrapolation, when posing counterfactual questions, and when pushing forward science.
This is the key difference between surrogate models in general and surrogate models of scientific value. 

For surrogate models of scientific value, we are at least as interested in the properties of an AI model itself as we are in its predictive performance, and we want to use the AI model to formulate testable hypotheses about the processes generating the data and the processes driving predictive performance. 
To know which counterfactuals to pose, it is important to have access to both the learned representation, $g_j$, and the decision function, $h_j$, that relates those representations to observable outcomes in the world. 
Studying data representations in the $g_j$ can reveal emergent system properties or controls that may themselves constitute discoveries; and understanding how these patterns interrelate through the application of $h_j$ can provides insight into system behaviors. 
Operationally, we aim to enable domain scientists to identify, test, and validate the properties of surrogate models, and their representations and decisions, thereby enabling the integration of data-driven AI models into the life-cycle of discovery in domain science, just as computational methodologies are fundamentally interwoven today. 

As a concrete example of a simple surrogate model and the AI model from which it could have been derived, consider how a Random Forest can be fitted to predict the area of a rectangle from data consisting the lengths of sides, $s_1$ and $s_2$, and areas, $A$, of rectangles. 
In this setting, the Random Forest will learn to approximate the response $A=s_{1}s_{2}$---at least within the domain of the training/testing data. 
Of course, when new data are provided, e.g., with longer/shorter side lengths than present in the training set, errors can be enormous (in this case, because Random Forests extrapolate with flat estimates of conditional expectations). 
If, however, via scientific counterfactual methods, one is able to 
develop a surrogate that captures the idea, or ``mechanism,'' 
that area is equal to the product of the lengths of the sides---a minimal first-principles model---then one expects the resulting surrogate to 
extrapolate to very different data far better than the original Random Forest model from which it was extracted. 
This is importantly different than ``model distillation,'' as the term is used currently in AI.
Typically, distillation aims to achieve similar prediction/generalization quality with a much smaller model, but it does not explicitly seek interpretability or explainability, as we use the terms \cite{hinton2015distilling}.
Developing technologies for scientifically validating surrogate models, in the presence of noise and complex local structure, e.g., for models that exhibit approximate local sparsity and/or intermediate-scale structure, is precisely the grand challenge we~champion.  

History can serve as a guide. 
The notion of a scientifically-validated surrogate model is not peculiar to bridging the gap between data-driven AI models and first-principles scientific models.
Recall Dirac's famous claim in 1929: ``The fundamental laws necessary for the mathematical treatment of a large part of physics and the whole of chemistry are thus completely known, and the difficulty lies only in the fact that application of these laws leads to equations that are too complex to be solved.''
This promise started to be delivered upon only with the development of surrogate models---Hartree-Fock models, correlated electron models, general energy models, etc.
These computationally-driven surrogate models, often called model chemistries, provide an approximate but well-defined theory that is consistent with---but not derived from---the underlying quantum mechanics.
They are scientifically validated, by scientific counterfactual methodologies, and their \emph{raison d'etre} is to provide qualitative interpretation of and predictive capability for chemical phenomenon~\cite{Pop99}.
These and related methods now form the basis for semi-empirical quantum chemistry, molecular mechanics simulations, and a large body of scientific and engineering work in solid state physics, chemical physics, and biophysics/biochemistry, including protein structure analysis, and drug discovery and~design. 


Understanding the context for the development of those surrogate models in computationally-enabled science will help to illuminate the path forward for the development of surrogate models for data-driven AI-enabled science. 
Of course, in chemical physics, we had understanding, but not efficient computability, and for AI models, we have efficient computability, but not understanding---the directions are reversed, but the bridge is the same: interpretable \emph{and} explainable surrogate models. 
Around and soon after the time of Dirac's claim (see Figure~\ref{fig:global-intermediate-local}C), analytical theory performed relatively-well for solid-state physics systems (which could be treated as an approximation of an ideal lattice) as well as for gas-state physical systems (which could be treated as an approximation of an ideal gas).
However, analytical theory largely failed for liquid-state systems (which are not meaningfully perturbative approximations of either of those analytically tractable cases) as well as for soft condensed matter and other much more complex chemical systems.
This, coupled with the availability of newly-inexpensive electronic computers, opened the door to fundamental developments in Monte Carlo, Markov Chain Monte Carlo, and Molecular Dynamics.
These developments, of course, have become much more widely applicable, and they are now the building blocks to solve a wide range of very practical scientific and engineering problems.

Revisiting Figure~\ref{fig:global-intermediate-local}, we believe that we are in an analogous situation.
We are told, e.g., that we are on the verge of general Artificial Intelligence, etc., etc, if only we had more data or more computation.
Currently, both traditional statistical theory as well as computationally-driven mathematical theory (e.g., theory for bootstrap methods or numerical simulation) perform reasonably well in relatively simple situations: with data that may be viewed as a perturbative approximation of some global or local statistical sparsity structure; or with a well-defined partial differential equation in a well-defined geometry, representing a physical model of the world.
However, both perform much more poorly when confronted with state-of-the-art AI models, which themselves are strongly-coupled learning machines, with properties depending strongly on the input data.
Simply asking for bigger, faster, better computers will not solve this problem.

The time is ripe for the prioritization of the development of \emph{data-driven scientifically-validated surrogate models of state-of-the-art AI models}.
In a manner analogous to how model chemistries are consistent with the principles of quantum mechanics, but capture chemical phenomenon in a quantitative way, these data-driven surrogate models will be consistent with the principles of statistical learning theory, and will capture, in detail, the behavior of the AI algorithms on which they are based.
They will be validated and evaluated by scientific counterfactual methods, not simply by proving a theorem or evaluating performance on training and testing data.
We expect that they too will be useful as building blocks for a broad range of very practical scientific and engineering problems.

To obtain from fitted AI models interpretable and explainable surrogates that are more amenable to scientific methods for hypothesis discovery will require the iterative feedback provided by hypothesis formation, experimentation, and testing, and refinement. 
An encouraging (and hopefully illustrative) example comes from recent work using symbolic regression to extract physical principles from the data representation learned by a Graph Neural Network during training \cite{cranmer2020discovering}. 
There, the model's internal data representation, $g$, was used to discover important system parameters---along with the structures of their interactions---and these parameters were fed into an algorithm to generate simple, low-dimensional equations that mimicked the decision function, $h$, of the Neural Network. 
Remarkably, the learned symbolic models often extrapolated better than the network from which they were derived---as we expect with first principles models. 
We highlight this example for it's methodology: patterns discovered by an AI algorithm become hypotheses, which are organized and prioritized through symbolic regression into testable models. 
This methodology goes beyond work on physics informed ML and physics informed Neural Networks \cite{raissi2018,raissi2019physics,BPK16,KGZKM21_TR}, 
work learning some fairly complex physics directly from low-dimensional data \cite{udrescu2020ai,liu2020ai}, and work showing that AI models can out-perform state-of-the-art physics-based models designed with deep domain knowledge~\cite{RCSx19,rutz2019atmospheric}. 
%

Importantly, this methodology also highlights how popular approaches to decoding Neural Networks are auxiliary or orthogonal to the task of learning scientifically-useful surrogates.
For example, saliency maps highlight specific features, but not their interactions, and they are not validated counterfactually.
In our notation, they provide some insight into the representation function, $g$, but they do not directly link $g$ to the decision rule $h$---they aim to provide the ``what'' but not the ``how.''
As such, they do not yet enable the discovery of first-principles understanding \cite{Adebayo2018, Kindermans2019}; and, in the absence of a counterfactual methodology, there is little reason to believe such maps are faithful, reliable, or informative representations of the world from which the data were derived \cite{janizek2020explaining, goebel2018explainable}. 
In a scientific sense, their insights are not \emph{testable}---and it is precisely the derivation of testable surrogate models that will lead to transformative advancements in science.

\section{Next steps}

Next steps will require the development of counterfactually-validated surrogate models for data-driven AI-enabled science that are as useful as model chemistries have been for computationally-enabled science.
This will require novel mathematical and statistical methods as well as a deep coupling between these theoretical techniques and scientific domain knowledge. 
\textit{However, outside of a few prototypical special cases, methods for extracting data representations and developing data-driven surrogate models do not exist, and few existing research programs are prioritizing their development.} 
Domain sciences tend to focus on domain science, treating essential enabling algorithmic and statistical methodologies as an afterthought.
To develop AI methods that will further scientific knowledge, \emph{we must develop methods to extract novel, testable hypotheses directly from data-driven AI models}, including prioritizing the following:
\begin{itemize}
    \item 
    Surrogate models derived from AI models: theory and methods for the extraction of first-principle data representations (surrogate models) from AI models that make explicit both the model's data representation ($g$) and its decision process ($h$).
    \item 
    Statistical analysis of surrogate models: statistical tests and metrics to enable hypothesis discovery and refinement, and the identification of interesting outliers.
    \item 
    Scientific analysis and scientific validation of surrogate models: strategies for counterfactual reasoning and experimental design---as a means to close the loop.
\end{itemize}
Progress on these directions will enable domain scientists to identify emergent system properties that govern behavior, captured by an AI model's internal data representation, $g$, and linked to system outputs through the learned decision function, $h$, in a scientifically falsifiable way.
The recent DOE AI Town Hall report identified a need for the creation of surrogate models for deep learning architectures as one of the essential capabilities that must be developed for the advancement of AI research in the next decade~\cite{stevens2020ai}.
In our view, developing procedures to obtain mechanistic insights from AI algorithms should be a top priority for the statistics and applied mathematics communities more generally \cite{Yu3920, Murdoch_2019, Kumbier_2018, stevens2020ai}.

As always, advancements will be driven by use-cases that give rise to the vanguard. 
As an example, in synthetic biology, the optimization of cell metabolism for bioproduction poses an enormous engineering challenge: to co-opt and jointly optimize the behavior of many thousands of metabolic processes that give rise to cell fitness to produce compounds that convey no survival benefit to the host on which they are imposed. 
Traditional synthetic biology approaches involve \emph{ad hoc} engineering practices predicated on human intuition, which lead to enormously long development times \cite{radivojevic2020machine}. 
Recently, high-throughput experimentation combined with the careful curation of data has enabled the AI-automation of some design tasks \cite{zhang2020combining}. 
However, uncertainty analysis remains challenging, leading to surprising model failures \cite{radivojevic2020machine}.
In addition, we lack the capacity to interrogate these fitted AI models to discover the emergent properties of cell metabolism that underlie predictive accuracy, e.g., in a manner analogous to how we interrogate complex models of protein structure that were developed with molecular dynamics and Monte Carlo techniques. 

If we accomplish this, then surrogate modeling strategies such as we propose have the potential to reveal new principles that govern cell metabolism---new \emph{knowledge} about how cells achieve dynamic homeostasis, learned by applying scientific methodologies to AI models and learning machines. 
To make this concrete, a hypothetical surrogate model for an AI model fitted to predict cell bioproduction from cell genetics might look as follows: a representation function, $g$, that links collections of genetic variants to metabolic fluxes for specific pathways; and a decision function, $h$, that maps interactions between metabolic fluxes or pathways to production levels of the target bioproduct.  
Such knowledge has eluded us in the study of individual metabolic pathways; and, to achieve robust control, it is essential to understand how the system works together, as a whole. 
In this and many other cases, the disconnect between predictive power and the generation of new fundamental hypotheses about how the world works resides in our failure to learn 
from AI models themselves.
Useful data-driven surrogate models should enable us to learn from AI models and to formulate scientific questions such that AI models---and their surrogates---can help us find the answers.

\section{Looking forward}

Any technology can be used for good or for ill.
However, models that are amenable to interrogation and interpretation have obvious advantages in increasingly-important societal applications, e.g., with ``ethical AI,'' going well beyond scientific and industry applications we have been discussing~\cite{Rud19}. 
A sobering reminder of the importance of this comes from the introduction of AI models into the criminal justice system, particularly for the prediction of the likelihood of recidivism, where blindly following the recommendations of AI models has already come under fire for providing potentially socially-biased predictions \cite{hao2019ai}. 
The cost of being wrong here is enormous---lasting or irreparable damage to lives and livelihoods. 
At present, solutions to understanding or measuring bias rely on humans in the loop, e.g., through crowdsourcing \cite{van2019crowdsourcing}. 
Human interpretable and explainable surrogates for data-driven AI models, as we propose, would permit more systematic analysis within social scientific frameworks, enabling the assurance of fairness using the same statistical frameworks we bring to bear in the justice system as it stands today. 

Similarly, medical institutes are trialing AI models within their diagnostic, prognostic, and therapeutic recommendation systems \cite{nagendran2020artificial}.
Here too, the cost of being wrong is far higher than for industry applications such as targeted advertising. 
Even within industrial applications, the recent failure of Amazon's AI-based decision support system for human resources and hiring practices underscores an important point: the lack of human understanding of how the AI model was making decisions made it impossible to repair, and the project was scrapped entirely \cite{amazon_sexist}. 
Similar problems occurred to Microsoft \cite{microsoft_racist_article, microsoft_racist_blog}, Apple \cite{apple_racist_article}, and Google \cite{google_racist_article}, among others. 
DARPA's Competency Aware Machine Learning (CAML) program seeks to provide uncertainty estimates, to develop learning machines that ``know when they don't know,'' for these and other applications.
So far, this program has focused on building learning machines that attempt to learn their own competency, or confidence, during model training. 
However, absent theoretical foundations and counterfactual methodologies such as those we have been advocating, there is little reason to trust their estimates of uncertainty any more than to trust their~predictions.

Scientifically-validated surrogate models, on the other hand, provide a framework which is largely orthogonal to the traditional methodology for AI model training.
As such, they provide the potential to identify and isolate vulnerabilities (e.g., due to Simpson's paradox \cite{wagner1982simpson, bickel1975sex, MM21a_simpsons_TR}) and to place measures of uncertainty for AI models on the more firm theoretical foundations already available for traditional statistical models and first-principles scientific models. 
This points to a ubiquitous challenge in data science: data are, by their nature, retrospective. 
The correlations leveraged for predictive accuracy may be falsely suggestive of causative mechanisms. 
This issue needs to be explored further in the context of AI decision support systems.
Such capabilities would address many potential concerns surrounding applications of AI, e.g., judicial procedures \cite{TechnicalFlawsOfPretrial_ML}, performance metrics for self-driving cars \cite{RR-2662}, drug discovery and prioritization \cite{schneider2018automating}, medical diagnostics \cite{neri2020artificial}, and many more. 

Certainly, the way in which data scientists and domain scientists interact with AI models will change in profound ways as we go forward.
Assume, for the moment, that the community is successful in developing the methods necessary to achieve what we have outlined. 
We will have enabled domain scientists to develop and test hypotheses, extract insight, and use modern, data-driven AI models to obtain scientific understanding in a way that is as seamlessly integrated into the cycle of scientific discovery as the computer and computational methodologies.
This advancement has the potential to be as transformative in the 21\textsuperscript{st} century as statistics was at the dawn of the 20\textsuperscript{th} century and as computer-aided simulation and modeling was at mid-20\textsuperscript{th} century.  
The relationships and patterns discovered through the application and analysis of AI models have the potential to become a discerning lens through which we view data, and thus through which we understand our world.
If we are successful, science in the 21\textsuperscript{st} century will be defined by our capacity to \textit{learn from learning machines}.

\section*{Acknowledgements}

We thank Susan Brand for her development and realization of the figures in this manuscript. We thank Meredith Murr and Melanie Mitchell for their contributions to the structure and presentation of this manuscript, and their many close readings and comments. We thank Aditi Krishnapriyan and Alejandro Queiruga for helpful comments and critiques that improved the quality of this manuscript.
\textbf{Funding:} JBB and DD were supported by the ExaSheds project, which was supported by the U.S. Department of Energy, Office of Science, Office of Biological and Environmental Research, Earth and Environmental Systems Sciences Division, Data Management Program, under Award Number DE-AC02-05CH11231. JBB was also supported by LBNL LDRD through contract DE-AC02-05CH1123, and National Science Foundation BIGDATA grant NSF-1613002. 
MWM would like to acknowledge DARPA, DOE, IARPA, NSF, and ONR for providing partial support of this work; our conclusions do not necessarily reflect the position or the policy of our sponsors, and no official endorsement should be inferred.
LPT, WBH, AWY, QJ, TC, and MA were supported by funding from Pattern Computer, Inc. 
HGM was supported by the Agile BioFoundry (http://agilebiofoundry.org) and the DOE Joint BioEnergy Institute (http://www.jbei.org), supported by the U. S. Department of
Energy, Energy Efficiency and Renewable Energy, Bioenergy Technologies Office, and the
Office of Science, through contract DE-AC02-05CH11231 between Lawrence Berkeley National Laboratory and the U.S. Department of Energy. HGM is also supported by the Basque Government through the BERC 2014-2017 program and by Spanish Ministry of Economy and Competitiveness MINECO: BCAM Severo Ochoa excellence accreditation SEV-2013-0323.
SP work was supported by the Director, Office of Science, Office of Advanced Scientific Computing Research, of the U.S. Department of Energy under Contract No. DE-AC02-05CH11231
GP and BLM acknowledge financial support from Laboratory Directed Research and Development (LDRD) program of the Los Alamos National Laboratory (LANL) via project \#20190001DR. LANL is operated by Triad National Security, LLC, for the National Nuclear Security Administration of U.S. Department of Energy, Contract No. 89233218CNA000001.
GVG acknowledges from support the NIHR Birmingham ECMC, NIHR Birmingham SRMRC, Nanocommons H2020-EU (731032) and the NIHR Birmingham Biomedical Research Centre and the MRC Heath Data Research UK (HDRUK/CFC/01), an initiative funded by UK Research and Innovation, Department of Health and Social Care (England) and the devolved administrations, and leading medical research charities. The views expressed in this publication are those of the authors and not necessarily those of the NHS, the National Institute for Health Research, the Medical Research Council or the Department of Health.
JM was supported by the U.S. Department of Energy, Office of Science, Office
of Advanced Scientific Computing Research, Applied Mathematics program under contract number DEAC02005CH11231.
SC was supported by National Institutes of Health Grant RF1-AG053303.
BJW was supported by Pacific Northwest National Laboratory is operated by Battelle Memorial Institute for the Department of Energy under contract DEAC05-76RLO1830.
DJ was supported by the Plant-Microbe Interfaces (PMI) and Secure Ecosystem Engineering and Design (SEED) Scientific Focus Areas in the Genomic Science Program as well as by The Center for Bioenergy Innovation (CBI). The U.S. Department of Energy Bioenergy Research Centers are supported by the Office of Biological and Environmental Research in the DOE Office of Science. DJ was also supported by the Exascale \& Petascale Networks for KBase and Integrated Pennycress Resilience projects funded by the Genomic Sciences Program from the U.S. Department of Energy, Office of Science, Office of Biological and Environmental Research.
SP was supported by the Artificial Intelligence Initiative, sponsored by the Laboratory Directed Research and Development Program of Oak Ridge National Laboratory, managed by UT-Battelle, LLC, for the U. S. Department of Energy.

\bibliography{xai}
\bibliographystyle{plain}

\end{document}